%% file: example.tex
\documentclass{article}

\usepackage{graphicx}
\usepackage{wrapfig}
\usepackage{caption}
\usepackage{subcaption}
\usepackage{booktabs}
\usepackage{array}
\usepackage{multirow}
\usepackage{siunitx}
\usepackage{amsmath}
\usepackage{amssymb}
\usepackage[printonlyused]{acronym}
\usepackage[noEnd=true]{algpseudocodex}
\usepackage{algorithm}
\usepackage[hidelinks]{hyperref}
\usepackage[bottom]{footmisc}
\input{acronyms}

\usepackage[preprint]{corl_2025}

\title{Learning Deployable Locomotion Control via Differentiable Simulation}

\author{
  Clemens Schwarke\textsuperscript{1,2}, Victor Klemm\textsuperscript{1}, Joshua Bagajo\textsuperscript{1}, \\ 
  \textbf{Jean-Pierre Sleiman\textsuperscript{1,3}, Ignat Georgiev\textsuperscript{4}, Jesus Tordesillas\textsuperscript{1,5}, Marco Hutter\textsuperscript{1}} \\
  \textsuperscript{1}ETH Zurich, \textsuperscript{2}NVIDIA, \textsuperscript{3}RAI Institute, 
  \textsuperscript{4}Georgia Institute of Technology, \\ \textsuperscript{5}Comillas Pontifical University\
}

\begin{document}
\maketitle

%===============================================================================

\begin{abstract}
    Differentiable simulators promise to improve sample efficiency in robot learning by providing analytic gradients of the system dynamics. Yet, their application to contact-rich tasks like locomotion is complicated by the inherently non-smooth nature of contact, impeding effective gradient-based optimization. Existing works thus often rely on soft contact models that provide smooth gradients but lack physical accuracy, constraining results to simulation. To address this limitation, we propose a differentiable contact model designed to provide informative gradients while maintaining high physical fidelity. We demonstrate the efficacy of our approach by training a quadrupedal locomotion policy within our differentiable simulator leveraging analytic gradients and successfully transferring the learned policy zero-shot to the real world. To the best of our knowledge, this represents the first successful sim-to-real transfer of a legged locomotion policy learned entirely within a differentiable simulator, establishing the feasibility of using differentiable simulation for real-world locomotion control.
        
\end{abstract}


%===============================================================================

\begin{figure}[htbp]
   \centering
   \includegraphics[width=0.82\textwidth]{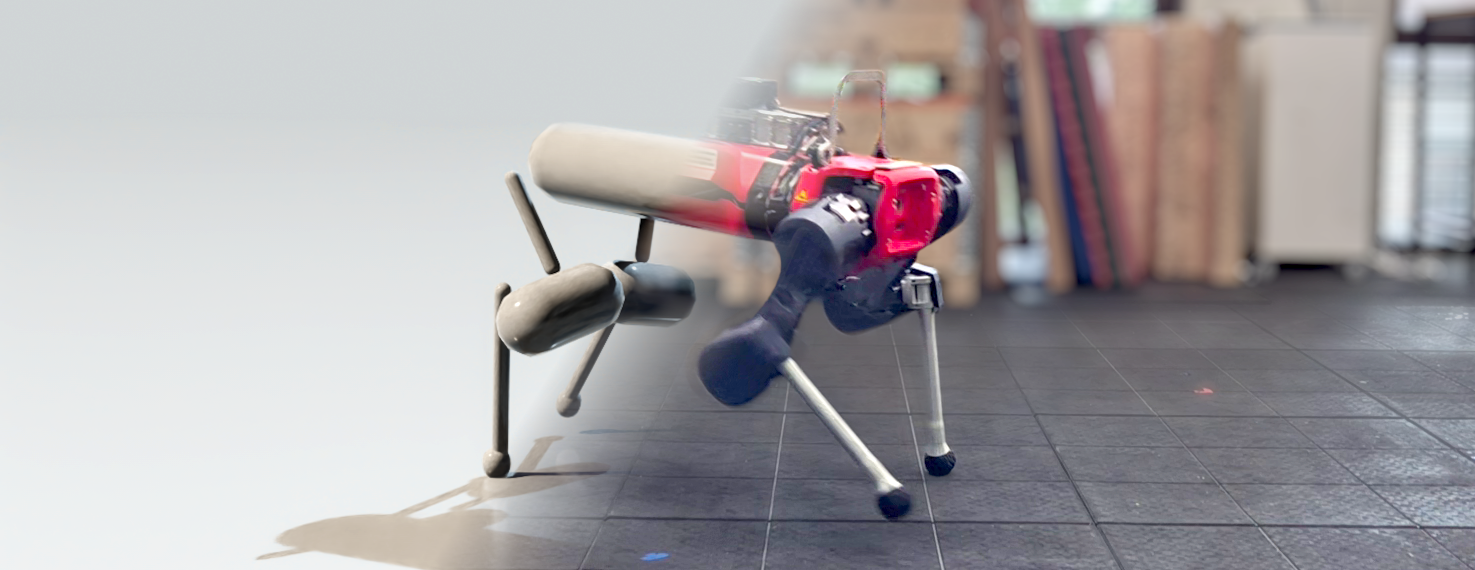}
   \captionsetup{width=.82\textwidth}
   \caption{A walking quadrupedal robot trained in a differentiable simulation. The supplementary video is available at: \href{https://youtu.be/UC4U4xn1e3w}{https://youtu.be/UC4U4xn1e3w}}
   \label{fig:title}
   \vspace{-0.1cm}
\end{figure}

\section{Introduction}
\label{sec:introduction}

\ac{RL} has emerged as a powerful framework for optimizing control policies that can solve a variety of robotic tasks, including legged locomotion~\cite{hwangbo2019learning, rudin2022learning}. Many \ac{RL} algorithms rely on sampling the task objective to estimate its gradient with respect to the policy parameters, often referred to as \ac{ZoG} estimation. This approach allows optimization even when the underlying system dynamics, such as those in conventional physics simulators, are non-differentiable. However, recent advancements in differentiable simulation frameworks enable the efficient computation of analytic \acp{FoG} of the system dynamics~\cite{hu2019, freeman2021, Werling2021, warp2022, howell2022}. These analytic gradients promise significantly improved sample efficiency and potentially better asymptotic performance compared to \ac{ZoG} estimates, due to their lower variance~\cite{Suh2022, wiedemann2022, georgiev2024adaptive}.

Despite their potential, leveraging \acp{FoG} effectively remains challenging, particularly for contact-rich robotic tasks such as legged locomotion \cite{hu2019, freeman2021, Degrave2019}. The core difficulty lies in the non-smooth and discontinuous nature of contact modeling, complicating the optimization landscape and hindering effective \ac{FoG}-based optimization. To mitigate the issue, many works employ soft contact models \cite{georgiev2024adaptive, xu2022, xing2024stabilizing, Geilinger2020}. While these models provide smooth, continuous gradients, they often sacrifice physical accuracy, confining results to simulation. Conversely, hard contact models offer high physical fidelity but yield discontinuous and less informative gradient signals, making \ac{FoG}-based optimization difficult. Consequently, learning robust locomotion policies purely within a differentiable simulator and successfully transferring them to the real world has remained an open challenge~\cite{freeman2021, georgiev2024adaptive, Degrave2019}.

To bridge this gap, we propose a differentiable contact model specifically designed to provide informative gradients for optimization while maintaining high physical fidelity suitable for real-world deployment. Inspired by the effects of stochastic smoothing observed in current \ac{RL} frameworks~\cite{Suh2022a}, our model is derived by analytically smoothing a hard contact formulation, akin to the approach in~\cite{Pang2023} for quasi-dynamic systems. To validate the benefits of our approach, we compare it against a soft contact model often used in recent works~\cite{warp2022, georgiev2024adaptive, xu2022, xing2024stabilizing} and a hard contact formulation~\cite{gehring2014evaluation,Carius2018}.

Applying our model, we train a quadrupedal locomotion policy entirely within our differentiable simulator using the \ac{SHAC} algorithm~\cite{xu2022}, utilizing analytic gradients derived from the proposed contact model. Crucially, we show that the learned policy transfers directly to a real quadrupedal robot. To the best of our knowledge, this represents the first successful sim-to-real transfer of a legged locomotion policy learned entirely within a differentiable simulator leveraging its analytic gradients. This result establishes the feasibility of using differentiable simulation for learning real-world locomotion control, paving the way for more sample-efficient learning paradigms in contact-rich robotic domains.

In summary, our main contributions are:
\begin{itemize}
\item A differentiable contact model that combines gradient informativeness and physical fidelity for contact-rich dynamics.
\item A demonstration of learning legged locomotion with high sample efficiency, using the analytic gradients of our differentiable simulation.
\item Successful zero-shot sim-to-real transfer of a learned locomotion behavior, validating the efficacy of the proposed contact formulation.
\end{itemize}

%===============================================================================

\section{Related Work}
\label{sec:related_work}

The benefits of optimizing with \acp{FoG} have been reported for various applications, such as soft body manipulation~\cite{Huang2021}, system identification from video~\cite{LeCleach2023, murthy2020gradsim}, or grasp synthesis~\cite{Turpin}. However, despite promising results, \ac{FoG}-based methods often encounter challenges related to complex and non-smooth optimization landscapes, which can result in diverging gradients and unstable learning~\cite{Suh2022}.

Several approaches have been proposed to mitigate these difficulties. To improve gradient quality in the presence of discontinuities, prior works have explored techniques such as smoothing collision geometries~\cite{Turpin}, employing ``leaky gradients'' to provide information in the absence of contact~\cite{Turpin}, and utilizing randomized smoothing or analytically smoothed dynamics~\cite{Suh2022a, Pang2023}. The concept of smoothing contact dynamics has also been explored in other areas, such as the field of model-predictive control~\cite{le2024fast,kim2023contact}. Some works address the issue of gradient divergence when differentiating long trajectories via \ac{BPTT} by training concurrent controllers~\cite{wiedemann2022} or optimizing actions at each time step~\cite{Mora2021}. Furthermore, hybrid methods combining \acp{FoG} and \acp{ZoG} have been investigated, using \acp{FoG} for sample generation~\cite{Qiao2021} or adaptively interpolating between gradient types~\cite{Suh2022}. Notably, the \ac{SHAC} algorithm~\cite{xu2022} successfully integrates \acp{FoG} into an actor-critic framework, demonstrating competitive performance on tasks like MuJoCo's ant locomotion~\cite{Todorov2012MuJoCoAP}.

Despite these advancements, applying \ac{FoG}-based learning to achieve physically realistic locomotion remains challenging~\cite{freeman2021, Degrave2019}. While simulation results using differentiable simulators seem promising~\cite{georgiev2024adaptive, xu2022, xing2024stabilizing}, their applicability to real-world robotics remains an open question due to potential discrepancies between simulated and real dynamics, especially concerning contact. Ultimately, the goal of learning robotic control is deployment on physical systems, making sim-to-real transfer a critical benchmark. To date, sim-to-real transfer of learned locomotion policies leveraging gradient-based optimization has been extremely limited. One notable success was recently achieved in~\cite{song2024learning}, however, they employed a non-differentiable simulator for forward simulation and a simplified single rigid-body dynamics model for gradient computation. This cleverly avoids the direct trade-off between gradient quality and physical realism within a single simulator but introduces the complexity of managing and aligning two separate models. Our work, in contrast, focuses on utilizing a single, fully differentiable simulation framework.

A key component enabling \ac{FoG}-based learning is the differentiable simulator itself. Numerous frameworks for differentiable rigid-body dynamics have been developed, differing in their differentiation methods, such as automatic differentiation~\cite{freeman2021, warp2022, Degrave2019}, symbolic differentiation~\cite{Werling2021}, or implicit differentiation~\cite{howell2022}, as well as their contact modeling approaches. For an introduction to contact modeling, we refer the reader to~\cite{horak2019similarities, le2024contact}. Generally, simulators employ either soft contact models~\cite{freeman2021,warp2022,Geilinger2020} which provide smooth gradients amenable to optimization but may lack physical accuracy, or hard contact models, often based on a \ac{LCP}~\cite{Werling2021} or a \ac{NCP}~\cite{howell2022}, which generally offer higher fidelity but yield less informative gradients. For legged locomotion, this trade-off has so far either prevented successful optimization~\cite{freeman2021,Degrave2019} or successful transfer to hardware. Our work addresses this gap by proposing a contact model designed to balance gradient informativeness with the physical fidelity required for sim-to-real transfer.

%===============================================================================

\section{Method}
\label{sec:method}

\begin{wrapfigure}{r}{0.45\textwidth}
\vspace{-1.5\intextsep}
\centering
\includegraphics[width=0.45\textwidth]{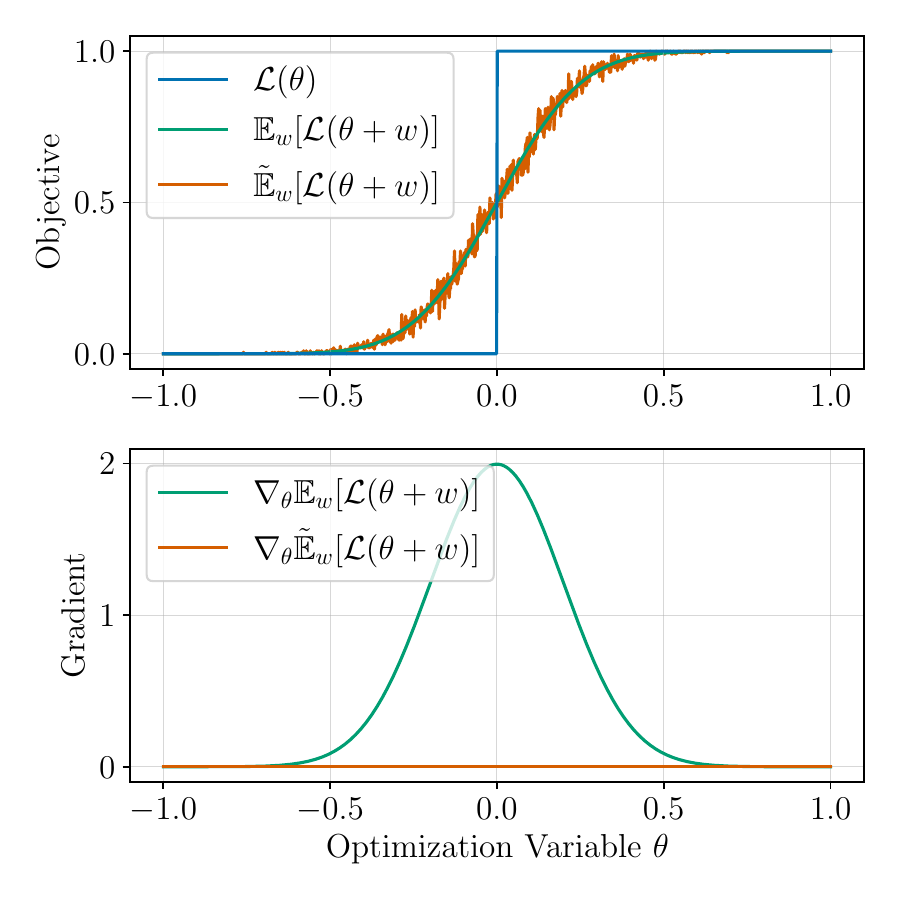}
\caption{An objective function \(\mathcal{L}(\theta)\) and its expected value under stochastic noise. While the approximation \(\tilde{\mathbb{E}}\) is an unbiased estimator of the true expected value, the gradient approximation of the expected value \(\nabla_{\theta}\tilde{\mathbb{E}}\) is biased if \(\mathcal{L}\) is discontinuous. A comprehensive analysis is given in \cite{Suh2022}.}
\vspace{-0.8cm}
\label{fig:riemann}
\end{wrapfigure}
Non-differentiable rigid-body simulators mainly employ hard contact models because they allow for large simulation time steps, do not require parameter tuning, and model physics with high fidelity. However, hard contact models introduce discontinuities into the dynamics and consequently into the optimization objective. 

\subsection{The Challenge with Discontinuities}
\label{sec:discontinuities}

To understand the effect of discontinuities on the optimization process, consider a generic optimization problem
\begin{equation}
    \min_{\theta} \mathcal{L}(\theta)\,,
\end{equation}
where \(\mathcal{L}\) is the objective or loss function and \(\theta\) is the optimization variable. \ac{ZoG}-based algorithms sample \(\mathcal{L}\) stochastically to construct a gradient estimate, effectively optimizing the expected value
\begin{equation}
      \mathbb{E}_w[\mathcal{L}(\theta+w)] = \int_w p(w)\mathcal{L}(\theta+w)\,dw \,,
\end{equation}
where \(w\) is stochastic noise and \(p(w)\) is its probability distribution. Under a finite number of samples, the expectation is approximated by
\begin{equation}
    \tilde{\mathbb{E}}_w[\mathcal{L}(\theta+w)]=\frac{1}{N}\sum_{n=0}^N \mathcal{L}(\theta+w_n)\,
\end{equation}
with \(w_n \sim p(w)\). Figure~\ref{fig:riemann} illustrates how introducing stochasticity effectively smooths discontinuities in the objective landscape, explaining the success of \ac{RL} methods even in such scenarios. Applying the same idea in the gradient domain yields
\begin{equation}
\label{eq:FoG}
    \nabla_{\theta}\tilde{\mathbb{E}}_w[\mathcal{L}(\theta+w)] = \frac{1}{N}\sum_{n=0}^N \nabla_{\theta} \mathcal{L}(\theta+w_n)\,,
\end{equation}
which does not accurately approximate the gradient of the stochastic objective in the presence of discontinuities. Instead, the sampled gradient estimate is biased, often yielding zero or misleading directions near discontinuities, as depicted in Fig.~\ref{fig:riemann}. The benefit of stochastic smoothing for \ac{ZoG} estimation does thus not directly transfer to \acp{FoG}. Even without discontinuities, stiff dynamics may lead to what~\cite{Suh2022} refers to as empirical bias, apparent under a small number of samples \(N\). In essence, the goal is to compute informative \acp{FoG}, without the need for sampling inherent to \ac{ZoG} estimation.

Optimizing with hard contact models faces an additional challenge stemming from discrete-time contact resolution. Most simulators detect contacts only at the beginning of each time step, allowing interpenetration between detection points. This discretization effect can alter or even invert the gradient direction compared to the underlying continuous-time dynamics, further complicating optimization \cite{hu2019}.

\subsection{Differentiable Rigid-Body Simulation}
\label{sec:diffsim}

To address these challenges, we implement a rigid-body simulation with a smooth, differentiable contact model within NVIDIA Warp~\cite{warp2022}. Warp is a Python-based kernel programming framework that facilitates high-performance computation through GPU parallelization and source code transformation to CUDA. It supports \acf{AD} by automatically generating adjoint kernels, rendering the implemented dynamics differentiable. Our simulation advances the system dynamics in generalized coordinates and is based on Moreau's time stepping scheme~\cite{moreau1988unilateral}, similar to~\cite{gehring2014evaluation,Carius2018}, which exhibits improved stability properties over the commonly used semi-implicit Euler integration~\cite{Bender2014}. A Gauss-Seidel algorithm resolves the \ac{NCP} of hard contact, which we adapt to achieve smooth dynamics. A detailed description of the simulation algorithm can be found in Appendix~\ref{sec:sim_details}.

\subsection{Analytic Smoothing of Contact Dynamics}
\label{sec:smoothing}

While stochasticity smoothes the discontinuities associated with hard contact, it does not provide meaningful \acp{FoG}, as previously presented in Sec.~\ref{sec:discontinuities}. Therefore, we propose to smooth the hard contact model analytically by substituting the discontinuous function of the contact force with respect to the penetration depth with a sigmoid function\footnote{While smoothing the function with Gaussian noise, often employed to introduce stochasticity in \ac{RL} methods, would yield the error function~\cite{Suh2022}, this work relies on the sigmoid function for its lower computational cost. The sigmoid function is equivalent to the step function smoothed with logistic noise~\cite{Pang2023}.}. Fundamentally, we achieve this by scaling the contact forces with
\begin{equation}
    \mathrm{sigmoid}(d,\kappa) = \frac{1}{1+e^{-d\kappa}} \,,
\end{equation}
where \(d\) is the penetration depth and \(\kappa\) determines the stiffness of the function.

Algorithm~\ref{alg:softgs} outlines the modified Gauss-Seidel scheme incorporating analytic smoothing. Inputs to the contact solver are the Delassus Matrix \(\boldsymbol{G}\), which expresses the system's inverse inertia in contact coordinates, \(\boldsymbol{c}\), which contains dynamic quantities that need to be counteracted by the contact impulses, and \(\boldsymbol{p}\), an initial guess for the impulses. The modified version additionally requires \(\boldsymbol{d}\), which contains the penetration depth for all active contacts, and the stiffness parameter \(\kappa\). The iteration count \(N\) is fixed to allow the solver loop to be unrolled during backpropagation~\cite{Carius2018}. The contact set \(C\) contains all potential contacts whose contact distance is below a certain threshold\footnote{The threshold should be chosen to limit the number of active contacts and thus computational cost, while allowing for forces at a distance for informative gradients. We set the threshold to \(\infty\) and only consider foot contacts.}. Further details about the general algorithm, such as the choice of the \(r\)-factor or the \(\mathrm{prox(\cdot)}\) operator, are provided in Appendix~\ref{sec:sim_details}.

\begin{wrapfigure}[17]{l}{0.5\textwidth}
\vspace{-2.0\intextsep}
\begin{minipage}{0.5\textwidth}
\begin{algorithm}[H]
\caption{Modified Gauss-Seidel Iteration}\label{alg:softgs}
\renewcommand{\algorithmicrequire}{\textbf{Input:}}
\renewcommand{\algorithmicensure}{\textbf{Output:}}
\begin{algorithmic}
\Require $\boldsymbol{G},\boldsymbol{c},\boldsymbol{p},\boldsymbol{d},\kappa$
\Ensure $\boldsymbol{p}$

\For{$N$ solver iterations}
    \For{contact $j\in C$}
        \State $\boldsymbol{s} \gets \boldsymbol{0}$
        \State $r \gets 0$
        \For{contact $k\in C$}
            \If{$j=k$}
                \State $\boldsymbol{s} \gets \boldsymbol{s} + \boldsymbol{G}_{jk}\boldsymbol{p}_k$
            \Else
                \State $\boldsymbol{s} \gets \boldsymbol{s} + \boldsymbol{G}_{jk}\boldsymbol{p}_k\cdot \mathrm{sigmoid}(d_k,\kappa)$
            \EndIf
            \State $r \gets r + \det(\boldsymbol{G}_{jk})$
        \EndFor
        \State $r \gets \frac{1}{1+r}$
        \State $\boldsymbol{p}_j \gets \mathrm{prox}(\boldsymbol{p}_j -r(\boldsymbol{s}+\boldsymbol{c}_j))$
    \EndFor
\EndFor

\State $\boldsymbol{p} \gets \boldsymbol{p}\cdot \mathrm{sigmoid}(\boldsymbol{d},\kappa)$

\end{algorithmic}
\end{algorithm}
\end{minipage}
\end{wrapfigure}

Analytic smoothing is incorporated via two distinct sigmoid scaling steps. Within each solver iteration, scaling the impulse of other contacts $k$ by $\mathrm{sigmoid}(d_k, \kappa)$ ensures that the calculation for contact $j$ accounts for the influence of contact $k$ based on the impulse magnitude it is expected to apply, given its penetration depth $d_k$. Once the solver iterations are complete, the final impulses $\boldsymbol{p}$ are scaled by $\mathrm{sigmoid}(\boldsymbol{d}, \kappa)$ to effectively apply the smoothing. This introduces contact forces at a distance, enabling gradient-based optimization to actively seek contact when desirable.

The level of smoothing may be incrementally reduced during the training process to shift towards a more precise model by adjusting \(\kappa\). The model can become arbitrarily stiff without destabilizing the simulation, as it converges to the hard contact case with growing \(\kappa\). However, we did not find such scheduling to be necessary for the studied task, as training and transfer were not sensitive to the level of smoothing.

\begin{figure}[b]
   \centering
   \includegraphics[width=\textwidth]{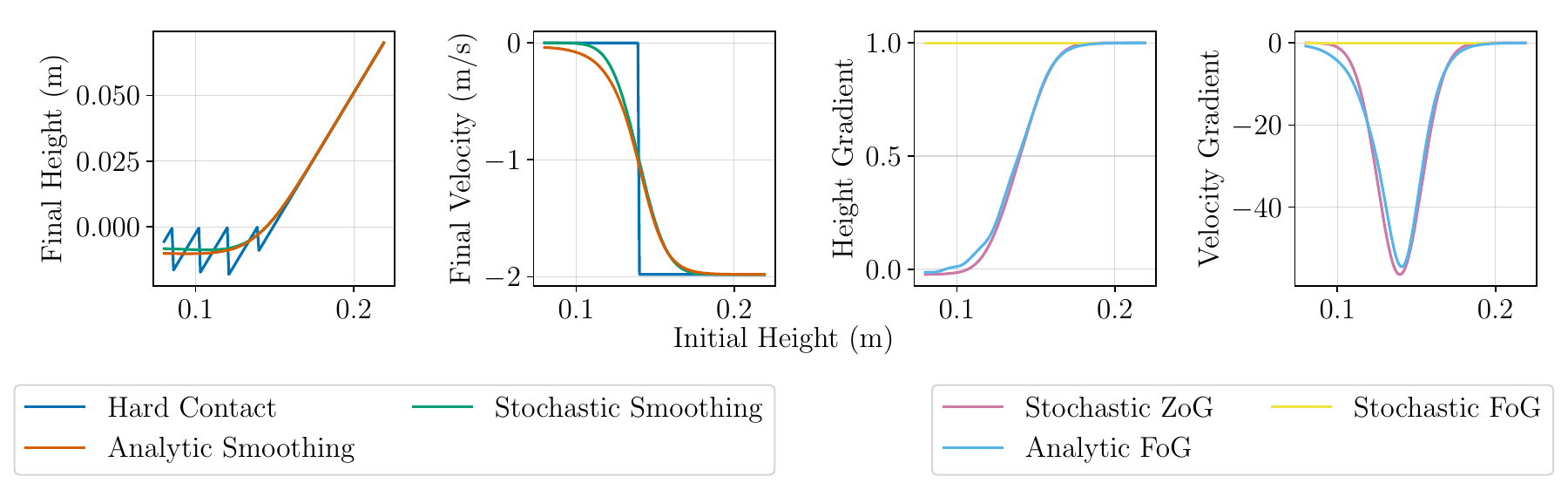}
   \caption{The final height, final velocity, and their gradients with respect to the initial height of a falling mass under gravity. The mass collides with the ground if the initial height is small enough (left half of each graph). Hard contact exhibits discontinuities in the position and velocity domains. Stochasticity smooths these discontinuities but yields a biased and uninformative \ac{FoG} gradient. The analytically smoothed contact model induces similar effects on the dynamics as stochasticity, with the advantage of informative and unbiased \acp{FoG}.}
   \label{fig:analytic_smoothing}
\end{figure}

Figure~\ref{fig:analytic_smoothing} demonstrates the effects of the proposed contact model on a toy example: an inelastic mass that is falling from different heights for a fixed time, potentially colliding with the ground. Hard contact exhibits two discontinuities. The first is visible as a sawtooth pattern in the mass's position, resulting from discrete time stepping and contact detection. The second discontinuity, evident in the mass's velocity, originates from the discontinuity of the normal contact force, which propagates through the dynamics and becomes apparent in the velocity domain. Analytic smoothing effectively mitigates these issues, closely replicating the smoothing effect of stochasticity, while providing informative and unbiased gradient signals for optimization.

\subsection{Considerations for Successful Learning and Sim-to-Real Transfer}
\label{sec:simtoreal}

Successfully learning legged locomotion and transferring it to physical hardware necessitates further consideration, particularly when training within a differentiable simulator. Key aspects include the choice of optimization algorithm, the fidelity of physical modeling, and strategies to enhance policy robustness.

First, directly optimizing over long task horizons using \acp{FoG} via \ac{BPTT} can lead to diverging and noisy gradients, especially in contact-rich scenarios~\cite{song2024learning, Mora2021}. To mitigate this, we employ the \ac{SHAC} algorithm~\cite{xu2022}. As a mixed-order method, \ac{SHAC} utilizes \acp{FoG} over a limited horizon and approximates the long-term return using a value function trained with \acp{ZoG}. This approach circumvents the challenges of \ac{BPTT} by restricting backpropagation to shorter, more manageable trajectory segments.

Second, the reward formulation needs to be fully differentiable. The formulation in~\cite{rudin2022learning} provides a suitable starting point for learning the quadrupedal velocity-tracking task but needs adaptation. For instance, we replace the non-differentiable feet air time reward, encouraging the robot to take large steps, with a differentiable reward that encourages the feet to follow a certain height trajectory, as detailed in Appendix~\ref{sec:anymal_setup}.

Third, achieving sim-to-real transfer requires accurate modeling of the actuator dynamics. A widely adopted approach is to model the dynamics with a learned actuator network~\cite{hwangbo2019learning}. However, integrating such networks would significantly increase the complexity of the differentiation graph required for \ac{FoG} computation, due to their recurrence to capture temporal dependencies. Therefore, we fit a simple \ac{PD} law to real-world actuator data~\cite{bjelonic2025sim2real}. This provides a reasonable approximation of the actuator dynamics while maintaining computational tractability for differentiation.

Finally, we incorporate domain randomization to account for unmodeled effects, such as sensor noise or force disturbances, to robustify the learned policy against the sim-to-real gap. Domain randomization further smoothes the optimization objective stochastically. A complete description of the environment setup, including the randomization and reward formulation, is provided in Appendix~\ref{sec:anymal_setup}.

%===============================================================================

\section{Experimental Results}
\label{sec:results}

We first train a quadrupedal robot to walk using a simple environment formulation, detailed in Appendix~\ref{sec:quadruped_setup}, similar to the MuJoCo ant environment studied in ~\cite{xu2022}. With this, we compare our method against the soft contact model used in~\cite{georgiev2024adaptive, xu2022, xing2024stabilizing} and the hard contact model obtained by not applying the proposed analytic smoothing. We also compare training with \ac{SHAC} against training with the purely \ac{ZoG}-based, state-of-the-art \ac{RL} algorithm \ac{PPO}~\cite{schulman2017proximal} to confirm the improvements in sample-efficiency presented in~\cite{georgiev2024adaptive,xu2022,xing2024stabilizing,song2024learning}. We then progress to training a velocity-tracking task with the quadrupedal robot ANYmal~\cite{hutter2016anymal}. 

\subsection{Comparison between Contact Models}
\label{sec:contactcomparison}

\begin{table}[ht]
    \centering
    \caption{The mean episode performance and standard deviation for different contact models.}
    \label{tab:rew}
    \sisetup{
      separate-uncertainty,
      detect-weight,
      reset-math-version = false,
    }
    \RenewDocumentCommand{\pm}{}{\mathbin{\mbox{\mathversion{normal}$\mathchar"2206$}}}
    \begin{tabular}{l
                    S[table-format=4.0(3)]
                    S[table-format=1.2(3)]
                    S[table-format=4.0(3)]
                    S[table-format=1.2(3)]}
        \toprule
        & \multicolumn{2}{c}{Evaluation with Training Model} & \multicolumn{2}{c}{Evaluation with Hard Contact} \\
        \cmidrule(lr){2-3} \cmidrule(lr){4-5}
        % Wrap column headers for S columns in braces {}
        {Training Model} & {Return} & {Episode Length (\unit{\second})} & {Return} & {Episode Length (\unit{\second})} \\
        \midrule

        Soft Contact
        & 2231 \pm 192
        & 9.69 \pm 0.28
        & 325 \pm 250
        & 1.58 \pm 1.15
        \\ \addlinespace

        Hard Contact
        & 2059 \pm 125
        & 9.57 \pm 0.15
        & 2059 \pm 125
        & 9.57 \pm 0.15
        \\ \addlinespace

        Smoothed Contact
        & \bfseries 2311 +- 62
        & \bfseries 9.88 +- 0.09
        & \bfseries 2255 +- 99
        & \bfseries 9.73 +- 0.25
        \\

        \bottomrule
    \end{tabular}
\end{table}

In the following, we examine whether different contact models enable learning locomotion and sim-to-sim transfer to an accurate hard contact model. Table~\ref{tab:rew} presents the final performance metrics for the locomotion task trained using \ac{SHAC}, comparing different contact models. The results were obtained by averaging the performance of $10$ training runs with different random seeds ($0$ to $9$) after $1000$ training iterations. Each evaluation involved simulating $100$ parallel environments for \qty{100}{\second}, with a maximum episode length of \qty{10}{\second}. All three contact models enable the agent to successfully learn the task, achieving high returns and episode lengths when evaluated within their respective training environments.

However, the models exhibit significant differences in their characteristics and transferability. The soft contact model suffers from two main issues. Firstly, smaller simulation steps are required to retain stability, reducing simulation speed by a factor of $2$ compared to the hard or smoothed contact simulations. Secondly, and more critically, policies trained with the soft contact model fail to transfer to the more physically realistic hard contact setting. This confirms that while the soft contact model facilitates efficient optimization~\cite{georgiev2024adaptive, xu2022, xing2024stabilizing}, its lack of physical accuracy prevents transfer to the real world. Increasing the contact stiffness to achieve higher realism destabilizes the gradient and prevents learning.

Conversely, the hard contact model provides physically accurate interactions. Interestingly, optimization with \ac{SHAC} is successful despite the inherent discontinuities in the contact dynamics. This suggests that the implicit smoothing provided by the value function of \ac{SHAC} helps navigate the challenging optimization landscape. Nevertheless, policies trained directly with hard contact exhibit unnatural walking behavior, involving suboptimal footholds and erratic movements, as can be seen in the supplementary video and the experiments in Appendix~\ref{sec:sup_results}. This may be attributed to misleading gradients arising from the contact discontinuities encountered during training.

\begin{wrapfigure}{r}{0.45\textwidth}
\vspace{-1.5\intextsep}
   \centering
   \includegraphics[width=0.45\textwidth]{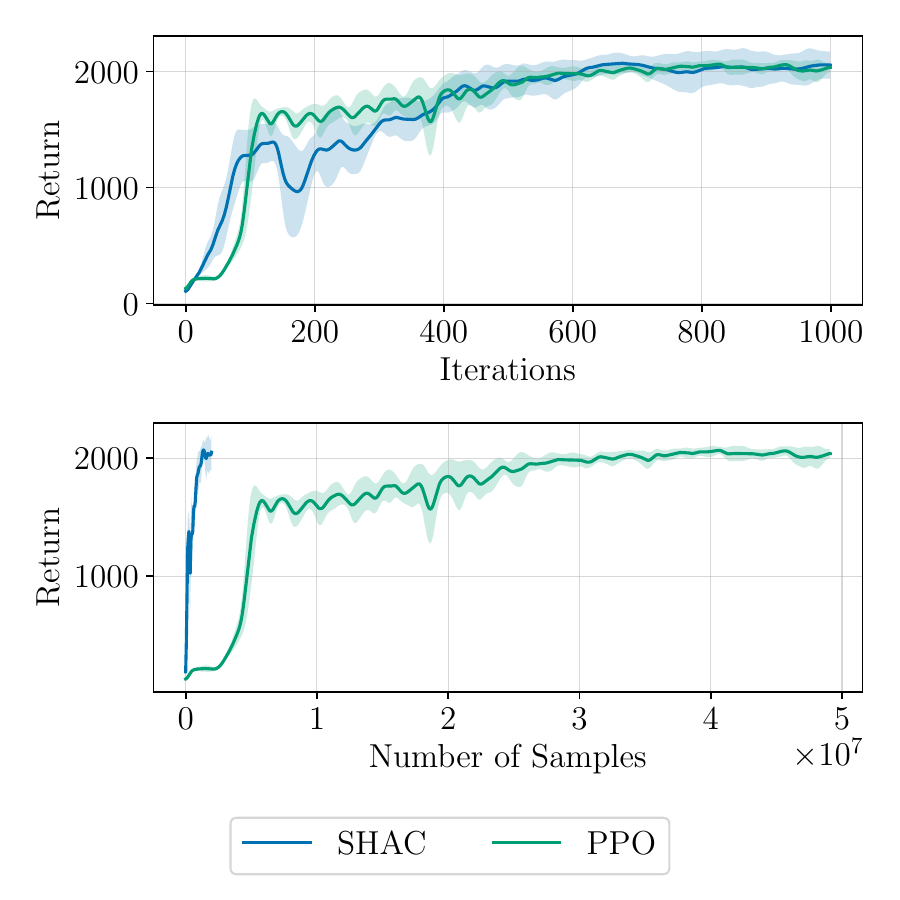}
   \caption{The episode return for the algorithms \ac{SHAC} and \ac{PPO} throughout training in terms of iterations and number of samples. \ac{SHAC} is trained with $64$ and \ac{PPO} with $2048$ parallel environments. The reward is averaged over five training runs with different random seeds ($0$ to $4$).}
   \label{fig:ppo_comparison}
\vspace{-1.2cm}
\end{wrapfigure}

The analytically smoothed contact model yields the highest performance when evaluated in its training environment, as well as when transferred to hard contact. As analytic smoothing is able to closely replicate the characteristics of stochastic smoothing, it might retain the underlying hard contact dynamics within its domain, explaining the successful transfer. While the approximately \qty{10}{\percent} increase in return compared to training with hard contact might appear modest, the qualitative difference in behavior is significant. The resulting locomotion behaviors are considerably smoother and show more natural and efficient gaits.

\subsection{Comparison between RL Algorithms}
\label{sec:algcomparison}

Next, we compare \ac{SHAC} to the publicly available \ac{PPO} implementation of RSL RL~\cite{rudin2022learning}. Both algorithms exhibit analogous convergence properties and achieve nearly identical final rewards, as shown in Fig.~\ref{fig:ppo_comparison}. Nevertheless, \ac{SHAC} outperforms \ac{PPO} in terms of sample efficiency by over an order of magnitude, owing to the reduced variance of \acp{ZoG}. However, \ac{PPO} demonstrates superior performance in terms of computational time with \qty{0.18}{\second} per iteration compared to \qty{0.33}{\second} with \ac{SHAC}, as detailed further in Appendix~\ref{sec:sup_results}. While \ac{SHAC} benefits from faster rollouts due to using fewer parallel environments, its learning phase is slower because it requires backpropagating through the entire rollout. This observation aligns with the findings in~\cite{xu2022} and suggests that the advantages of \ac{FoG}-based optimization become more apparent for higher-dimensional problems, such as humanoid control or learning from vision~\cite{luo2024residual}.

\subsection{Training a Velocity-Tracking Policy for Real-World Deployment}
\label{sec:deployment}

\begin{wrapfigure}{l}{0.45\textwidth}
\vspace{-2.0\intextsep}
   \centering
   \includegraphics[width=0.45\textwidth]{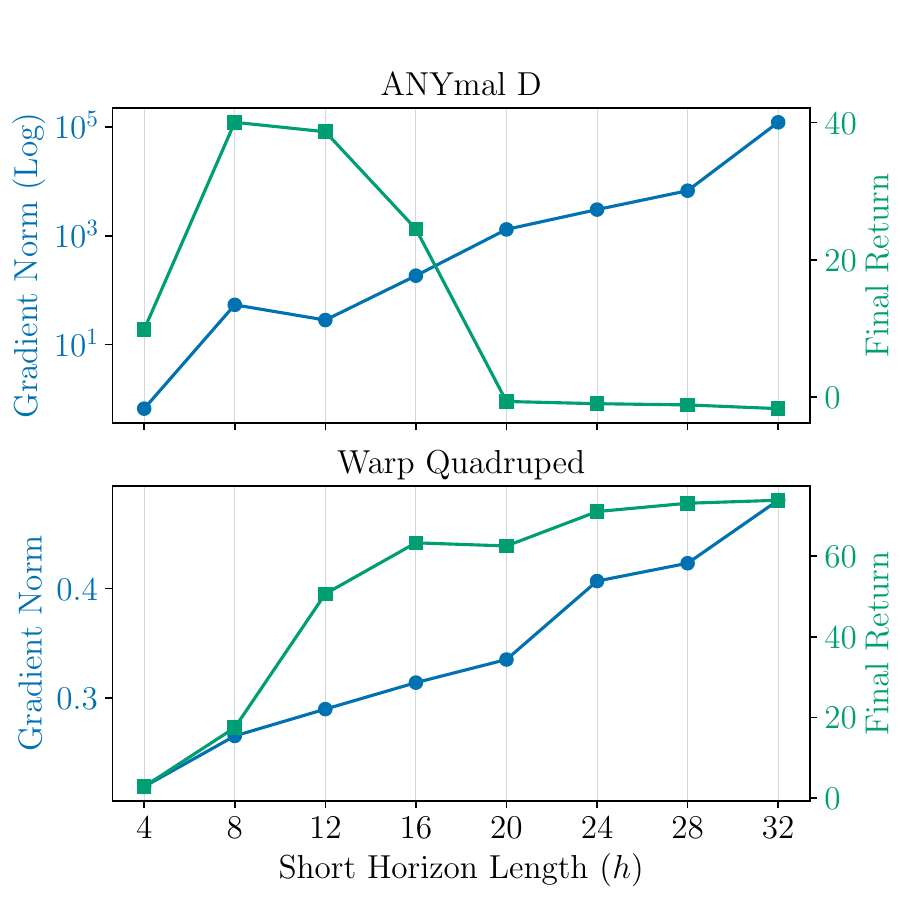}
   \caption{The mean gradient norm during training of the velocity-tracking task  and the final return after $1000$ iterations across different horizon lengths for both robots.}
   \label{fig:grad_norm}
\vspace{-0.5cm}
\end{wrapfigure}

Moving from the quadruped provided with Warp to the ANYmal robot proves challenging, due to its mass distribution and inertia. While the previously used quadruped features a light base and high leg inertia, ANYmal has a heavy base and light shanks with low inertia, resulting in significantly less smooth dynamics. Furthermore, ANYmal's actuators deliver four times the maximum torque. The implications on the learning process can be seen in Fig.~\ref{fig:grad_norm}. While the mean gradient norm during training with Warp's quadruped remains well below $1.0$ for any horizon length, the gradient norm with ANYmal increases exponentially and prohibits successful learning when backpropagating \acp{FoG} over more than $16$ steps. This highlights another major mismatch between simulation benchmarks, such as the ant environment~\cite{Todorov2012MuJoCoAP}, and real physical systems, further compared in Appendix~\ref{sec:sup_results}. We use a short horizon length of $12$ to facilitate stable optimization.

\begin{wrapfigure}{r}{0.45\textwidth}
\vspace{-1.5\intextsep}
   \centering
   \includegraphics[width=0.45\textwidth]{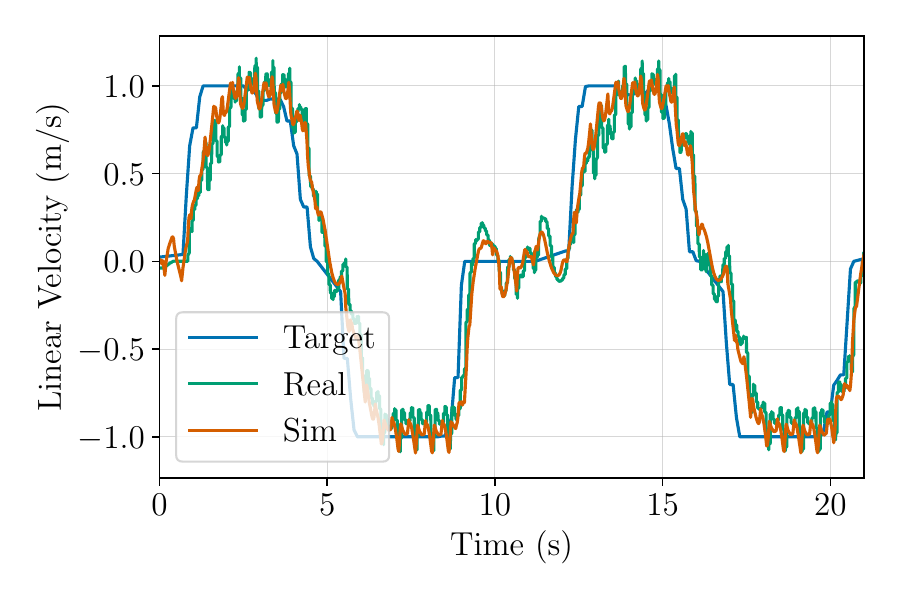}
   \caption{The target velocity for walking forward/backward and the measured velocity on the real robot and in simulation.}
   \label{fig:vel_tracking}
\vspace{-0.5cm}
\end{wrapfigure}
After training, the learned locomotion skill transfers to the real robot zero-shot. The robot is able to closely follow the velocity commands given by the operator as shown in Fig.~\ref{fig:vel_tracking}. Notably, the real velocity trajectory closely matches the one recorded in our differentiable simulation, confirming its sim-to-real capabilities. The learned control policy demonstrates high levels of robustness, withstanding force perturbations, as shown in the accompanying video. In summary, our experiments confirm the successful learning, zero-shot transfer, and robust real-world execution of locomotion policies trained using the proposed method.

%===============================================================================

\section{Conclusion and Future Work}
\label{sec:conclusion}

This work addressed the challenge of leveraging analytic gradients for the contact-rich task of legged locomotion by introducing an analytically smoothed contact model that balances informative gradients with the physical fidelity needed for hardware deployment. We confirmed the advantages of analytic smoothing by comparing our model to two alternative contact modeling approaches. We then successfully transferred a locomotion skill, learned entirely within our differentiable simulator, zero-shot to a real quadrupedal robot. This validates the feasibility of \ac{FoG}-based optimization even for complex, contact-rich tasks, offering a path towards more sample-efficient robot learning.

In future work, it would be interesting to scale this approach to systems with high-dimensional action spaces, like humanoids, or high-dimensional observation spaces, such as images, to leverage the improved sample efficiency. Lastly, incorporating differentiable rough terrain would make this approach more competitive with current \ac{RL} methods.

%===============================================================================

\section{Limitations}
\label{sec:limitations}

While our approach enables successful sim-to-real transfer, several limitations remain. \ac{FoG}-based optimization with \ac{SHAC}, although sample-efficient, incurs higher wall-clock time per iteration than optimization with \ac{PPO} due to the added need for differentiation. This computational overhead might negate the gained benefits of higher sample efficiency in low-dimensional optimization problems. Furthermore, to maintain computational tractability, we used a simple \ac{PD} law for actuator modeling rather than a more complex learned network. This simplification might reduce simulation fidelity, potentially affecting transfer for tasks demanding high actuator precision. Finally, the need for full differentiability imposes a constraint not present in \ac{ZoG} methods, requiring more effort in aspects such as simulation design or reward engineering.

%===============================================================================

\acknowledgments{This project was supported by the Swiss National Science Foundation through the National Centre of Competence in Automation (NCCR automation) and received funding from the European Union’s Horizon Europe Framework Programme under grant agreement No. 101070596.}

%===============================================================================

\bibliography{example}  % .bib

\clearpage

%===============================================================================

\appendix

\section*{Appendix}
\renewcommand{\thesubsection}{A.\arabic{subsection}}%

In the following, we provide additional details referenced in the main text, covering the simulation framework, experimental setups, and supporting results.

\subsection{Simulation Implementation}
\label{sec:sim_details}

This section describes the implemented simulation routine to provide deeper intuition into the contact resolution algorithm and the simulation in general. We also detail the soft contact formulation used in~\cite{warp2022, georgiev2024adaptive, xu2022, xing2024stabilizing}.

The fundamental principle of simulating a system of rigid bodies forward in time is the integration of the system's \ac{EoM} to find future velocities and positions. The \ac{EoM} can be stated in a general form as
\begin{equation}
    \boldsymbol{H}(\boldsymbol{q})\ddot{\boldsymbol{q}} + \boldsymbol{h}(\boldsymbol{q},\dot{\boldsymbol{q}}) = \boldsymbol{\tau} + \boldsymbol{J}_c^\mathsf{T}\boldsymbol{f}_c\,,
\end{equation}
where \(\boldsymbol{q}\) describes the position, \(\dot{\boldsymbol{q}}\) the velocity\footnote{The velocity is generally not equal to the time derivative of the generalized coordinate vector \(\boldsymbol{q}\). However, to align with the literature, the velocity term is still referred to as \(\dot{\boldsymbol{q}}\), despite the slight abuse of notation.}, and \(\ddot{\boldsymbol{q}}\) the acceleration of the system in generalized coordinates. \(\boldsymbol{H}(\boldsymbol{q})\) denotes the inertia matrix, \(\boldsymbol{h}(\boldsymbol{q},\dot{\boldsymbol{q}})\) includes centrifugal, Coriolis, and gravitational forces, \(\boldsymbol{\tau}\) is the actuation of the system, and \(\boldsymbol{f}_c\) includes all contact forces and is projected onto the generalized coordinate space by the contact Jacobian \(\boldsymbol{J}_c\). Note that the \ac{EoM} may also be formulated in maximal coordinates. However, many robotics simulators rely on generalized coordinates to reduce the number of necessary system constraints. 

Since analytic integration is generally intractable, numerical integration schemes approximate the continuous-time behavior of the system. First-order integrators, such as Euler's method, are often chosen for their simplicity. We implement Moreau's Time Stepping scheme~\cite{moreau1988unilateral}, a more accurate integration method also discussed in~\cite{gehring2014evaluation, Carius2018}. As detailed in Alg.~\ref{alg:moreau}, this scheme begins with an explicit Euler half-step to find the system's position at the midpoint of the simulation step \(i\) with length \(h\).

Subsequently, the simulator evaluates the individual \ac{EoM} terms to calculate the acceleration required for integration. First, \(\boldsymbol{h}(\boldsymbol{q},\dot{\boldsymbol{q}})\) is found by assuming all accelerations to be zero and then solving the inverse dynamics problem, i.e., finding the forces that lead to a given motion. The \ac{RNEA} efficiently computes the inverse dynamics for kinematic trees~\cite{featherstone2014rigid}. Second, \(\boldsymbol{H}\) can be constructed using the \ac{CRBA}~\cite{featherstone2014rigid}. Arguably, the most challenging task is finding the appropriate contact forces \(\boldsymbol{f}_c\).

\begin{algorithm}
\caption{Moreau's Time Stepping Scheme}\label{alg:moreau}
\renewcommand{\algorithmicrequire}{\textbf{Input:}}
\renewcommand{\algorithmicensure}{\textbf{Output:}}
\begin{algorithmic}
\Require $\boldsymbol{q}_i,\dot{\boldsymbol{q}}_i, \boldsymbol{\tau}_i$
\Ensure $\boldsymbol{q}_{i+1},\dot{\boldsymbol{q}}_{i+1}$
\State $\boldsymbol{q}_{\text{mid}} \gets \boldsymbol{q}_i + \frac{h}{2}\dot{\boldsymbol{q}}_i$
\Comment{Explicit Euler half-step}

\State $\boldsymbol{h} \gets \text{\ac{RNEA}}(\boldsymbol{q}_{\text{mid}},\dot{\boldsymbol{q}}_i, \boldsymbol{\tau}_i)$\Comment{\ac{RNEA} merges control inputs into \(\boldsymbol{h}\)}
\State $\boldsymbol{H} \gets \text{\ac{CRBA}}(\boldsymbol{q}_{\text{mid}})$

\State$ \boldsymbol{J}_c,\boldsymbol{G},\boldsymbol{c},\boldsymbol{p}
 \gets \textsc{toContact}(\boldsymbol{H},\boldsymbol{h},\boldsymbol{q}_{\text{mid}},\dot{\boldsymbol{q}}_i)$

\For{$N$ solver iterations}\Comment{Gauss-Seidel iteration}
    \For{$j = 1, \dots, n_c$}\Comment{Iterate over $n_c$ active contacts}
        \State $\boldsymbol{s} \gets \boldsymbol{0}$
        \State $r \gets 0$
        \For{$k = 1, \dots, n_c$}
            \State $\boldsymbol{s} \gets \boldsymbol{s} + \boldsymbol{G}_{jk}\boldsymbol{p}_k$
            \State $r \gets r + \det(\boldsymbol{G}_{jk})$
        \EndFor
        \State $r \gets \frac{1}{1+r}$
        \State $\boldsymbol{p}_j \gets \text{prox}(\boldsymbol{p}_j -r(\boldsymbol{s}+\boldsymbol{c}_j))$
    \EndFor
\EndFor

\State $\dot{\boldsymbol{q}}_{i+1} \gets  \dot{\boldsymbol{q}}_{i} + \boldsymbol{H}^{-1}(\boldsymbol{J}_c^\mathsf{T}\boldsymbol{p}-h\boldsymbol{h})$ \Comment{Midpoint step}
\State $\boldsymbol{q}_{i+1} \gets  \boldsymbol{q}_{i} + \frac{h}{2}(\dot{\boldsymbol{q}}_{i}+\dot{\boldsymbol{q}}_{i+1})$

\end{algorithmic}
\end{algorithm}

Instead of computing the forces and integrating them over the time step, the contact solver directly determines the required impulses \(\boldsymbol{p}\) to satisfy all contact constraints at the end of the simulation step. The first part of the procedure, summarized in Alg.~\ref{alg:moreau} as \(\textsc{toContact}(\cdot)\), involves projecting parts of the \ac{EoM} onto the contact domain using the contact Jacobian \(\boldsymbol{J}_c\). This Jacobian relates variations in the generalized coordinates \(\boldsymbol{q}\) with variations in the local contact coordinates. Note that \(\boldsymbol{J}_c\) includes all active contacts of the system, i.e.,
\begin{equation}
    \boldsymbol{J}_c = 
    \begin{bmatrix} 
    \boldsymbol{J}_{1} \\
    \boldsymbol{J}_{2} \\
    \vdots \\
    \boldsymbol{J}_{n_c} \\
    \end{bmatrix}\,,
\end{equation}
where \(n_c\) is the number of active contacts. Given the spatial Jacobian \(\boldsymbol{J}_s\), a byproduct of the \ac{CRBA} algorithm, the contact Jacobians for each active contact \(j\) can be readily computed by
\begin{equation}
    \boldsymbol{J}_j = \boldsymbol{J}_{s_k,P} - [\boldsymbol{r}]_{\times} \boldsymbol{J}_{s_k,R}\,,
\end{equation}
where index \(k\) selects the part of \(\boldsymbol{J}_s\) that corresponds to the rigid body at which the contact occurs, indices \(P\) and \(R\) denote the positional and rotational part of the Jacobian, and \(\boldsymbol{r}\) is a vector from the rigid body's origin to the contact point.

The first result of the projection using \(\boldsymbol{J_c}\) is the Delassus Matrix \(\boldsymbol{G}\), which expresses the system's inverse inertia in the contact coordinates. It is found using
\begin{equation}
    \boldsymbol{G} = \boldsymbol{J}_c \boldsymbol{H}^{-1} \boldsymbol{J}_c^\mathsf{T}\,.
\end{equation}
However, computing the inverse inertia \(\boldsymbol{H}^{-1}\) explicitly is generally avoided due to its computational cost and potential numerical instability. Therefore, \(\boldsymbol{G}\) is computed more efficiently by utilizing the factors of \(\boldsymbol{H}\), e.g., from reordered Cholesky factorization. The next term required is the vector \(\boldsymbol{c}\), computed with
\begin{equation}
\label{eq:c}
    \boldsymbol{c} = \boldsymbol{J}_c (\dot{\boldsymbol{q}}_i + h\boldsymbol{H}^{-1}\boldsymbol{h})\,,
\end{equation}
which includes dynamic quantities that need to be counteracted by the contact impulses to achieve \(\boldsymbol{v}_c=0\) at the time step's end. 

All contact impulses are initialized to
\begin{equation}
\label{eq:init}
    \boldsymbol{p}_j = -\boldsymbol{G}_{jj}^{-1}\boldsymbol{c}_j\,,
\end{equation}
where \(\boldsymbol{p}_j\) and \(\boldsymbol{c}_j\) represent the \(j\text{-th}\) elements of \(\boldsymbol{p}\) and \(\boldsymbol{c}\), corresponding to the \(j\text{-th}\) active contact. In contrast to~\cite{Carius2018}, tangential impulses are not set to zero, which yields more accurate results for a limited number of iterations in this work.

Initializing \(\boldsymbol{p}\) according to Eq.~\ref{eq:init} does not yet account for the interactions between contacts and the constraints of the friction cone. Thus, the second part of the contact handling procedure involves iterating over all contacts in a Gauss-Seidel fashion to converge to appropriate impulses considering interactions between contacts. In this approach, the impulses for each contact are updated sequentially using the most recently computed values of all other impulses. Every impulse update aims to reduce the current constraint violation at the corresponding contact. The update for \(\boldsymbol{p}_j\), before enforcing the friction cone constraint, can be expressed as
\begin{equation}
    \boldsymbol{p}_{j,\text{update}} = \boldsymbol{p}_j - r_j \, (\sum_{k=1}^{n_c} \boldsymbol{G}_{jk}\boldsymbol{p}_k + \boldsymbol{c}_j)\,.
\end{equation}
The convergence properties of the iteration scheme are influenced by the \(r\)-factor, and several strategies for choosing an appropriate value have been proposed~\cite{Andrews2022}. In this work, \(r\) is chosen to be
\begin{equation}
    r_j = \frac{1}{1+\sum_{k=1}^{n_c}|\boldsymbol{G}_{jk}|}\,,
\end{equation}
a minor variation from~\cite{Carius2018}, to ensure that \(r\) is always bounded. Lastly, a proximal projection~\cite{parikh2014proximal,studer2009numerics} with the proximal operator
\begin{equation}
    \text{prox}(\boldsymbol{p}_j) = \begin{cases}
        \boldsymbol{0}, & \text{if } p_{n_j} \leq 0 \\
        \boldsymbol{p}_j, & \text{if } p_{n_j} > 0 \wedge \|\boldsymbol{p}_{t_j}\| \leq \mu p_{n_j} \\
        [p_{n_j}, \mu p_{n_j} \frac{\boldsymbol{p}_{t_j}^\mathsf{T}}{\|\boldsymbol{p}_{t_j}\|}]^\mathsf{T}, & \text{if } p_{n_j} > 0 \wedge \|\boldsymbol{p}_{t_j}\| > \mu p_{n_j} \\
    \end{cases}
\end{equation}
at each update step ensures that the contact forces always remain within the bounds of the friction cone.

Once the iterative solver has determined the final contact impulses \(\boldsymbol{p}\) after \(N\) iterations,  the generalized velocity \(\dot{\boldsymbol{q}}_{i+1}\) can be computed using
\begin{equation}
\label{eq:velocity_update}
\dot{\boldsymbol{q}}_{i+1} = \dot{\boldsymbol{q}}_{i} + \boldsymbol{H}^{-1}(\boldsymbol{J}_c^\mathsf{T}\boldsymbol{p} - h\boldsymbol{h})\,.
\end{equation}
As with the computation of \(\boldsymbol{G}\) and \(\boldsymbol{c}\), multiplication by the inverse inertia is performed efficiently using the factorization of \(\boldsymbol{H}\) rather than explicit inversion. Finally, the generalized positions are updated using a trapezoidal rule, averaging the velocities at the beginning and end of the step with
\begin{equation}
\label{eq:position_update}
\boldsymbol{q}_{i+1} = \boldsymbol{q}_{i} + \frac{h}{2}(\dot{\boldsymbol{q}}_{i}+\dot{\boldsymbol{q}}_{i+1})\,.
\end{equation}
This completes simulation step \(i\), yielding the state \((\boldsymbol{q}_{i+1}, \dot{\boldsymbol{q}}_{i+1})\) needed for the next iteration.

Implementing this routine in Warp~\cite{warp2022} allows us to obtain gradients via \ac{AD}. While Warp handles differentiation of the core dynamics, the iterative nature of the Gauss-Seidel contact solver requires special consideration. By fixing the number of iterations, we make the solver amenable to \ac{AD}, as the loop can be unrolled during backpropagation. This method offers considerable implementation simplicity compared to analytical alternatives based on the implicit function theorem~\cite{howell2022,le2024fast, Belbute-Peres2018}. Furthermore, it bypasses the need to derive and implement complex analytical sensitivities, e.g., of the contact Jacobian or inverse inertia. Consequently, for small iteration counts, $N=10$ in this work, differentiating the unrolled solver within Warp presents a practical and potentially more efficient strategy for obtaining gradients through the contact dynamics in our framework.

The soft contact model used as a baseline in this work does not rely on an iterative solver. Instead, normal contact forces can be directly computed using
\begin{equation}
    f_n = \begin{cases} k_pd-k_d\min(v_n, 0), & \text{if } d \geq 0 \\ 
    0, & \text{otherwise} \end{cases}
\end{equation}
and tangential friction forces follow
\begin{equation}
\label{eq:fric_contact_soft}
\boldsymbol{f}_t = - \frac{\boldsymbol{v}_t}{\left\|\boldsymbol{v}_t\right\|_2} \min(k_f\left\|\boldsymbol{v}_t\right\|_2,\mu f_n)\,,
\end{equation}
where \(\boldsymbol{v}\) is the relative velocity in the contact frame, and \(\mu\) is the friction coefficient. The parameters \(k_p\), \(k_d\), and \(k_f\) have to be tuned and trade off physical accuracy against simulation stability and gradient smoothness.

\subsection{Environment Setup for Quadrupedal Locomotion}
\label{sec:quadruped_setup}

The experiments in Sec.~\ref{sec:contactcomparison} and Sec.~\ref{sec:algcomparison} were conducted using a simple environment, closely related to the one studied in~\cite{xu2022} for the ant robot, for better comparability and reproducibility. However, we train the quadruped provided with Warp~\cite{warp2022} to walk forward with a constant velocity of \qty{1.0}{\meter/\second} to better align with the goal of achieving deployable locomotion control. Table~\ref{tab:rew_simple} specifies the used rewards. The observations received by the control policy are identical to~\cite{xu2022}, but listed in Tab.~\ref{tab:obs_simple} for completeness. The policy outputs position commands for the joints that are then tracked with a \ac{PD} law. The maximum torque applied by the actuators is \qty{20}{\newton m}. Environments are terminated if the center of the quadruped's base falls below \qty{0.25}{m} with respect to the ground or if the maximum episode length of \qty{10}{s} is reached. All training parameters are listed in Tab.~\ref{tab:params_simple}.

\subsection{Environment Setup for Velocity-Tracking with ANYmal}
\label{sec:anymal_setup}

The environment formulation for the velocity-tracking task differs from the previously discussed setup in its reward and observation formulations, detailed in Tab.~\ref{tab:rew_track} and Tab.~\ref{tab:obs_track}, respectively. As mentioned in the main text, we modify the reward formulation of~\cite{rudin2022learning} to ensure differentiability. The original formulation encourages large steps via long foot swing durations, which aids transfer to real terrain but is inherently non-differentiable. We replace this reward with a differentiable reward for tracking a foot height reference. Although the task is learnable without the reward, it significantly enhances transferability to ground that is not perfectly flat. Note that this reward can be removed when training on rough terrain. The height reference is generated using a sinusoid, phase-shifted by \(\pi\) for alternating feet, and can be expressed as
\begin{equation}
    z_{foot}^* = 0.1\cdot \max(\sin(4 \pi t),0)\,,
\end{equation}
where \(t\) is the current time in seconds.

Additionally, domain randomization is introduced to facilitate sim-to-real transfer. This includes adding noise to observations, varying the robot mass and friction coefficient across environments, and changing the robot's base velocity at randomly sampled time steps. Table~\ref{tab:params_track} specifies the randomization ranges from which values are uniformly sampled. Furthermore, each environment receives a sampled velocity command that the robot is tasked with tracking. Lastly, to better approximate the real-world actuator dynamics in simulation, we fit the parameters of the \ac{PD} law for joint position target tracking and the joint armature using real-world data~\cite{bjelonic2025sim2real}.

\subsection{Additional Experimental Results}
\label{sec:sup_results}

This section supplements the main experimental results with additional data on computational performance during training, robot characteristics relevant for well behaved gradients, and examplary joint trajectories during locomotion.

Table~\ref{tab:timing_comparison} provides a breakdown of the computational time per training iteration for the algorithms \ac{SHAC} and \ac{PPO}. All experiments in this work were conducted on a single NVIDIA RTX 2080 Ti graphics card and the reported times are averaged over $5$ training runs. While \ac{SHAC} requires significantly fewer samples per iteration compared to \ac{PPO}, its learning phase takes longer due to the need for backpropagation through the simulation rollout. This results in a longer total time per iteration for \ac{SHAC} compared to \ac{PPO}, despite faster rollouts. Thus, the benefit of \ac{FoG}-based methods is expected to become relevant in higher-dimensional problem settings, where sample generation is a bottleneck.

\begin{table}[htbp]
\centering
\caption{The computational time for one training iteration.}
\label{tab:timing_comparison}
    \begin{tabular}{lcccc}
        \toprule
        Algorithm & Rollout Time (\unit{\second}) & Learning Time (\unit{\second}) & Total Time (\unit{\second}) & Samples\\
        \midrule
        SHAC & $0.086 \pm 0.009$ & $0.244 \pm 0.013$ & $0.330 \pm 0.019$ & $2048$ \\
        \addlinespace
        PPO  & $0.130 \pm 0.009$  & $0.053 \pm 0.005$ & $0.183 \pm 0.012$ & $49152$ \\
        \bottomrule
    \end{tabular}
\end{table}

Table~\ref{tab:robot_properties} compares physical properties of the ANYmal D robot used in our final experiments with the Warp Quadruped and the commonly used MuJoCo Ant benchmark. While all three robots are of comparable size, they exhibit substantial differences in total weight and in the ratio of base to leg mass. ANYmal's base to shank mass ratio is over an order of magnitude higher compared to the other two robots. Combined with its higher maximum actuator torque, this leads to substantially less smooth dynamics, as discussed in Sec.~\ref{sec:deployment} and reflected in the gradient norm analysis in Fig.~\ref{fig:grad_norm}. These differences underscore the importance of evaluating new methods on physically realistic models and, ultimately, on real robots.

\begin{table}[htbp]
\centering
\caption{Comparison of Robot Properties.}
\label{tab:robot_properties}
    \begin{tabular}{l
                    S[table-format=1.2]
                    S[table-format=2.2]
                    S[table-format=1.2]
                    S[table-format=2.0]
                    }
        \toprule
        Robot & {Height (\unit{\meter})} & {Torso Mass (\unit{\kilogram})} & {Shank Mass (\unit{\kilogram})} & {Torque (\unit{\newton\meter})} \\
        \midrule
        MuJoCo Ant~\cite{Todorov2012MuJoCoAP} & 0.75 & 0.33 & 0.10 & 1 \\
        \addlinespace
        Warp Quadruped~\cite{warp2022} & 0.58 & 6.20 & 2.00 & 20 \\
        \addlinespace
        ANYmal D & 0.70 & 25.00 & 0.76 & 80 \\
        \bottomrule
    \end{tabular}
\end{table}

\begin{figure}[htbp]
   \centering
   \begin{subfigure}[h]{0.45\textwidth}
    \includegraphics[width=\textwidth, trim=1cm 2cm 0cm 2cm, clip]{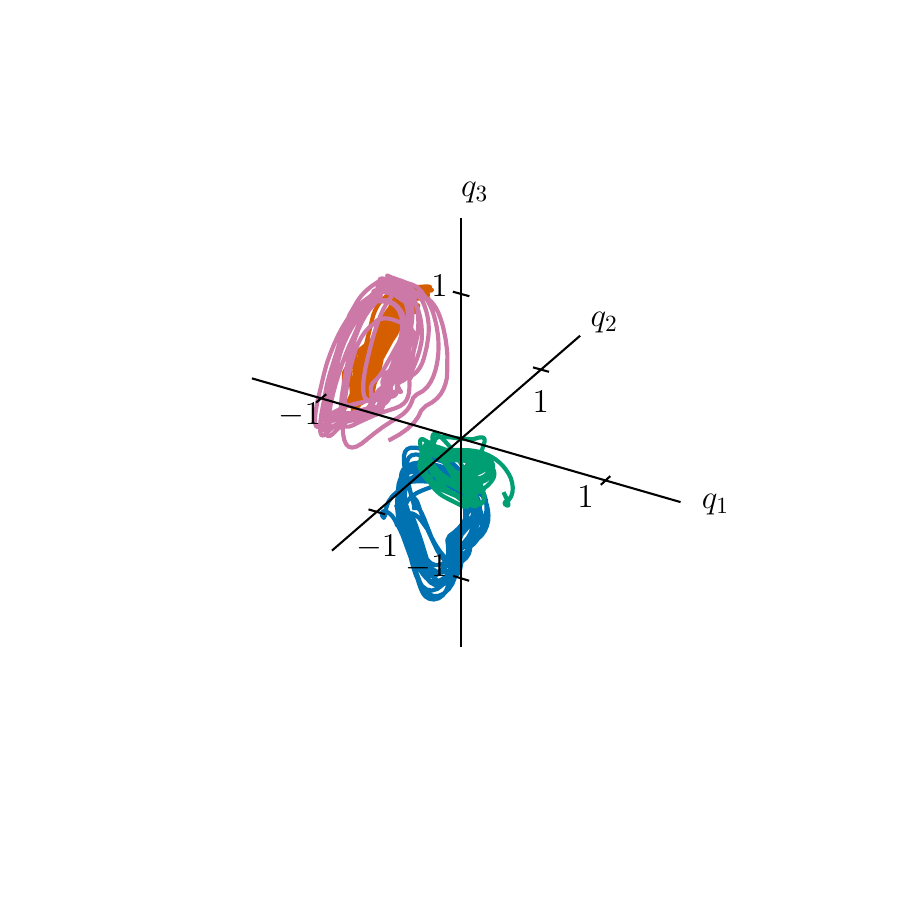}
   \end{subfigure}
   \begin{subfigure}[h]{0.45\textwidth}
    \includegraphics[width=\textwidth, trim=1cm 2cm 0cm 2cm, clip]{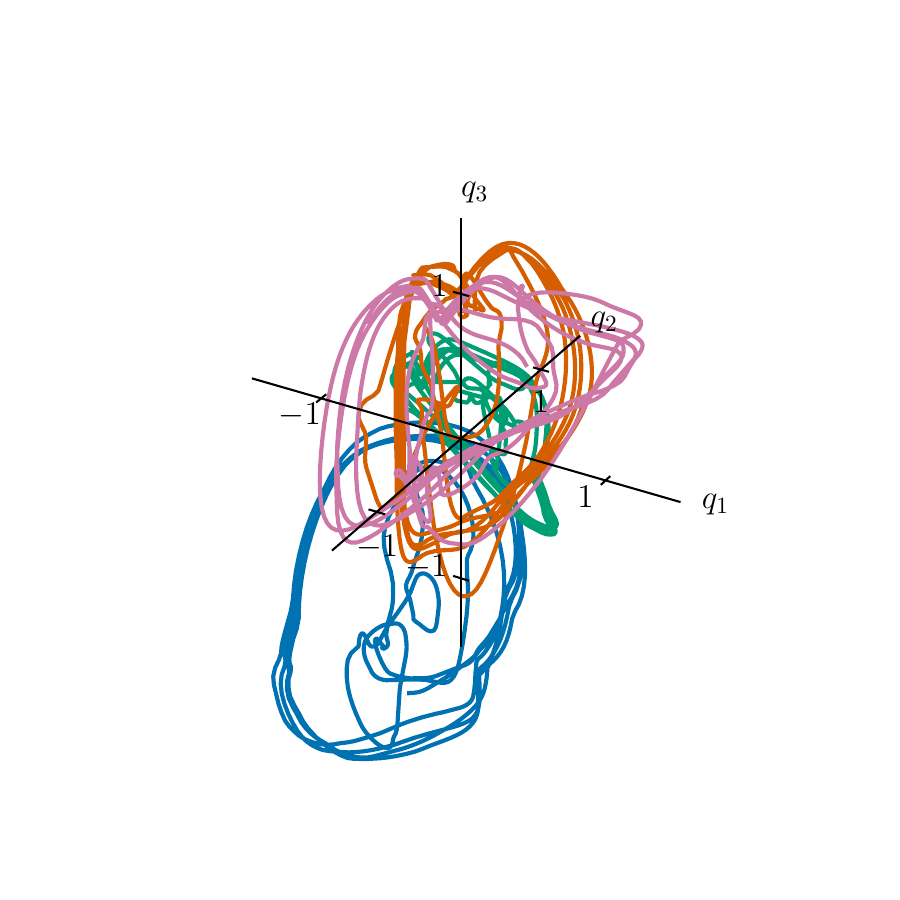}
   \end{subfigure}
   \caption{The joint position trajectories of the four legs (shown in different colors) of the quadruped robot, comparing a policy trained with smoothed contact (left) to a policy trained with hard contact (right). Both policies are evaluated using hard contact.}
   \label{fig:joints_quad}
\end{figure}

Figure~\ref{fig:joints_quad} highlights the difference in motion quality between policies trained with smoothed and hard contact models. The policy trained with smoothed contact exhibits smooth, periodic joint trajectories. In contrast, the policy trained with hard contact produces a less regular gait. Furthermore, the smoothed contact policy utilizes a significantly smaller range of joint motion, resulting in a more efficient motion that contributes to the higher return presented in Tab.~\ref{tab:rew}.

\begin{figure}[htbp]
   \centering
   \includegraphics[width=\textwidth]{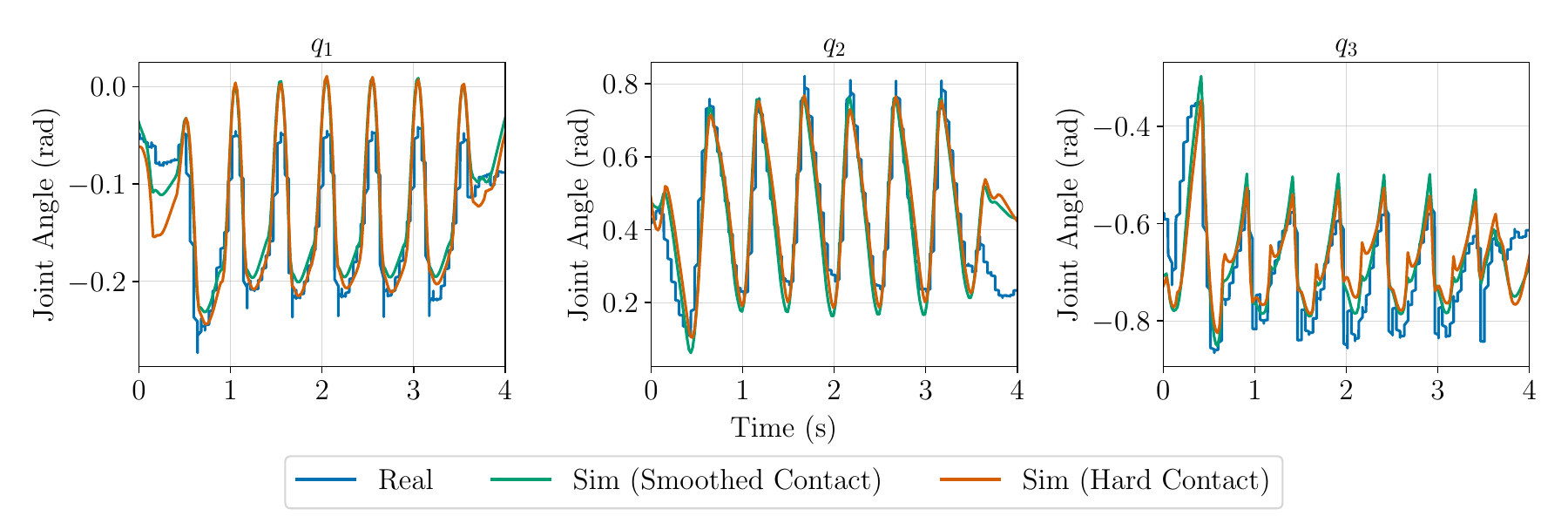}
   \caption{The joint position trajectories of ANYmal's left front leg, comparing real-world execution with simulation using smoothed and hard contact.}
   \label{fig:joints}
\end{figure}

Figure~\ref{fig:joints} shows joint position trajectories for the ANYmal robot during locomotion, comparing real-world execution with simulations using both smoothed and hard contact. The trajectories from the simulation closely match the real-world data, validating the successful sim-to-real transfer. Furthermore, the trajectories from the smoothed contact simulation are highly similar to those from the hard contact simulation. This strong agreement suggests that our smoothed contact model captures the essential dynamics of hard contact and that the remaining sim-to-real gap mainly originates from factors beyond contact modeling, such as unmodeled actuator dynamics or system parameters. The smooth and periodic nature of all trajectories reflects the stable gait learned by the policy.

%===============================================================================

\begin{table}[htbp]
\begin{center}
 \caption{The rewards for the quadruped locomotion task.}
 \label{tab:rew_simple}
    \begin{tabular}{lll} 
        \toprule
            Name & Formula & Weight\\
        \midrule
            Base velocity & \(\exp(-|v_x- 1.0|)\) & \(1.0\)  \\ 
            Base height & \(\exp(-|z-0.45|)\) & \(0.5\)  \\ 
            Base alignment & \(\boldsymbol{e}_{z}^{\text{world}} \cdot \boldsymbol{e}_{z}^{\text{base}}\) & \(0.5\)  \\ 
            Action magnitude & \(\sum_i\exp(-|a_i|)\) & \(0.01\)  \\ 
            Joint velocity & \(-\|\dot{\boldsymbol{q}}\|^2\) & \(0.001\)  \\
        \bottomrule
    \end{tabular}
\end{center}
\end{table}

\begin{table}[htbp]
\begin{center}
 \caption{The observations for the quadruped locomotion task.}
 \label{tab:obs_simple}
    \begin{tabular}{ll} 
        \toprule
            Name & Symbol \\
        \midrule
            Base height & \(z\in\mathbb{R}^{1}\) \\
            Base orientation & \(\boldsymbol{\xi}\in\mathbb{R}^{4}\) \\
            Linear base velocity & \({}_\mathcal{B}\boldsymbol{v}\in\mathbb{R}^{3}\) \\ 
            Angular base velocity & \({}_\mathcal{B}\boldsymbol{\omega}\in\mathbb{R}^{3}\) \\ 
            Joint position & \(\boldsymbol{q}\in\mathbb{R}^{12}\) \\ 
            Joint velocity & \(\dot{\boldsymbol{q}}\in\mathbb{R}^{12}\) \\ 
            Base alignment & \(\boldsymbol{e}_{z}^{\text{world}} \cdot \boldsymbol{e}_{z}^{\text{base}}\in\mathbb{R}^{1}\) \\
            Base heading alignment & \(\boldsymbol{e}_{x}^{\text{world}} \cdot \boldsymbol{e}_{x}^{\text{base}}\in\mathbb{R}^{1}\) \\
            Previous action & \(\boldsymbol{a}_{\text{prev}}\in\mathbb{R}^{12}\) \\
        \bottomrule
    \end{tabular}
\end{center}
\end{table}

\begin{table}[htbp]
\begin{center}
 \caption{The parameters for the quadruped locomotion task.}
 \label{tab:params_simple}
    \begin{tabular}{ll} 
        \toprule
            Name & Value\\
        \midrule
            \multicolumn{2}{c}{Environment} \\
        \midrule
            Episode length & \qty{10.0}{\second} \\
            Environment time step length & \qty{0.01}{\second} \\   
        \midrule
            \multicolumn{2}{c}{Simulation} \\
        \midrule
            Simulation time step length & \qty{0.01}{\second} \\
            Soft simulation time step length & \qty{0.0005}{\second} \\
            Gauss-Seidel iterations & $10$ \\
            Smoothing parameter $\kappa$ & $300$ \\
            Friction coefficient & $0.8$ \\
            Joint stiffness & $20.0$ \\
            Joint damping & $1.0$ \\
            Soft contact stiffness & $1.2 \cdot 10^4$ \\
            Soft contact damping & $3.0 \cdot 10^1$ \\
            Soft contact friction & $9.0 \cdot 10^2$ \\
        \midrule
            \multicolumn{2}{c}{\ac{SHAC}} \\
        \midrule
            Training iterations & $1000$ \\
            Parallel environments & $64$ \\
            Short horizon length & $32$ \\
            Actor learning rate & $0.002$ \\
            Critic learning rate & $0.002$ \\
            Actor learning rate decay & $0.995$ \\
            Critic learning rate decay & $0.997$ \\
            Discount factor & $0.99$ \\
            Value estimation & $0.95$ \\
            Target value network & $0.2$ \\
            Critic iterations & $16$ \\
            Critic mini batches & $4$ \\
            Maximum actor gradient norm & $1.0$ \\
            Maximum critic gradient norm & $10.0$ \\
        \midrule
            \multicolumn{2}{c}{\ac{PPO}} \\
        \midrule
            Training iterations & $1000$ \\
            Parallel environments & $2048$ \\
            Rollout length & $24$ \\
            Learning rate & $0.001$ \\
            Discount factor & $0.99$ \\
            Value estimation & $0.95$ \\
            Value loss coefficient & $1.0$ \\
            Entropy coefficient & $0.0$ \\
            Learning epochs & $5$ \\
            Mini batches & $4$ \\
            Desired KL divergence & $0.01$ \\
            Maximum gradient norm & $1.0$ \\
        \midrule
            \multicolumn{2}{c}{Network architecture} \\
        \midrule
            Actor MLP dimensions & $(128,64,32)$ \\
            Critic MLP dimensions & $(64,64)$ \\
            Activation function & ELU \\
        \bottomrule
    \end{tabular}
\end{center}
\end{table}

\begin{table}[htbp]
\begin{center} \caption{The rewards for the velocity-tracking task.}
 \label{tab:rew_track}
    \begin{tabular}{lll} 
        \toprule
            Name & Formula & Weight\\
        \midrule           
            Linear velocity tracking & \( \exp(-\frac{1}{0.25} \| {}_\mathcal{B}\boldsymbol{v}_{xy} - {}_\mathcal{B}\boldsymbol{v}_{xy}^* \|^2 ) \) & \(1.0\)  \\ 
            Angular velocity tracking & \( \exp(-\frac{1}{0.25} ( {}_\mathcal{B}{\omega}_{z} - {}_\mathcal{B}{\omega}_{z}^* )^2 ) \) & \(0.5\) \\
            Foot height tracking & \(\sum_{j} \frac{z_{foot,j}^*}{0.1} \cdot \exp(-\frac{1}{0.05} ( z_{foot,j} - z_{foot,j}^* )^2 )\) & \(3.0\)  \\ 
            Linear velocity error & \( - {}_\mathcal{B}v_z^2 \) & \(2.0\)  \\ 
            Angular velocity error & \( - \| {}_\mathcal{B}\boldsymbol{\omega}_{xy} \|^2 \) & \(0.05\)  \\ 
            Base height & \( \exp(-\frac{1}{0.1}(z - 0.45)^2) \) & \(1.0\)  \\ 
            Base orientation & \( - \| {}_\mathcal{B}\boldsymbol{g}_{xy} \|^2 \) & \(0.5\)  \\ 
            Action magnitude & \( - \sum_i |a_i| \) & \(0.05\)  \\ 
            Action rate & \( - \| \boldsymbol{a} - \boldsymbol{a}_{\text{prev}} \|^2 \) & \(0.01\)  \\ 
            Joint acceleration &  \( - \| \ddot{\boldsymbol{q}} \|^2 \) & \(2.5\cdot 10^{-7}\)  \\ 
            Joint torque & \( - \| \boldsymbol{\tau} \|^2 \) & \(2.5\cdot 10^{-5}\)  \\ 
        \bottomrule
    \end{tabular}
\end{center}
\end{table}

\begin{table}[htbp]
\begin{center}
 \caption{The observations for the velocity-tracking task.}
 \label{tab:obs_track}
    \begin{tabular}{ll} 
        \toprule
            Name & Symbol \\
        \midrule
            Linear base velocity & \({}_\mathcal{B}\boldsymbol{v}\in\mathbb{R}^{3}\) \\ 
            Angular base velocity & \({}_\mathcal{B}\boldsymbol{\omega}\in\mathbb{R}^{3}\) \\ 
            Projected gravity & \({}_\mathcal{B}\boldsymbol{g}\in\mathbb{R}^{3}\) \\
            Velocity command & \(({}_\mathcal{B}\boldsymbol{v}_{xy}^{*},{}_\mathcal{B}\omega_z^{*})^\top\in\mathbb{R}^{3}\) \\
            Joint position & \(\boldsymbol{q}\in\mathbb{R}^{12}\) \\ 
            Joint velocity & \(\dot{\boldsymbol{q}}\in\mathbb{R}^{12}\) \\ 
            Previous action & \(\boldsymbol{a}_{\text{prev}}\in\mathbb{R}^{12}\) \\
            Phase & \(\sin(4\pi t)\in\mathbb{R}^1\)\\
        \bottomrule
    \end{tabular}
\end{center}
\end{table}

\begin{table}[htbp]
\begin{center}
 \caption{The parameters for the velocity-tracking task.}
 \label{tab:params_track}
    \begin{tabular}{ll} 
        \toprule
            Name & Value\\
        \midrule
            \multicolumn{2}{c}{Environment} \\
        \midrule
            Episode length & \qty{20.0}{\second} \\
            Environment time step length & \qty{0.02}{\second} \\
        \midrule
            \multicolumn{2}{c}{Simulation} \\
        \midrule
            Simulation time step length & \qty{0.005}{\second} \\
            Gauss-Seidel iterations & $10$ \\
            Smoothing parameter $\kappa$ & $300$ \\
            Joint stiffness & $85.0$ \\
            Joint damping & $0.6$ \\
            Joint armature & $0.1$ \\
        \midrule
            \multicolumn{2}{c}{\ac{SHAC}} \\
        \midrule
            Training iterations & $20000$ \\
            Parallel environments & $128$ \\
            Short horizon length & $12$ \\
            Actor start learning rate & $0.005$ \\
            Critic start learning rate & $0.002$ \\
            Actor final learning rate & $1.0 \cdot 10^{-5}$ \\
            Critic final learning rate & $1.0 \cdot 10^{-5}$ \\
            Discount factor & $0.99$ \\
            Value estimation & $0.95$ \\
            Critic iterations & $16$ \\
            Critic mini batches & $4$ \\
            Maximum actor gradient norm & $1.0$ \\
            Maximum critic gradient norm & $10.0$ \\
        \midrule
            \multicolumn{2}{c}{Randomization} \\
        \midrule
            Friction coefficient range & $[0.5,1.25]$ \\
            Added base mass range & $[\qty{-5.0}{\kilogram},\qty{5.0}{\kilogram}]$ \\
            Random base velocity range & $[\qty{-0.5}{\meter/\second},\qty{0.5}{\meter/\second}]$ \\
            Random base velocity interval range & $[\qty{10.0}{\second},\qty{15.0}{\second}]$ \\
            Velocity command range for ${}_\mathcal{B}\boldsymbol{v}_{xy}^{*}$ & $[\qty{-1.0}{\meter/\second},\qty{1.0}{\meter/\second}]$ \\
            Velocity command range for ${}_\mathcal{B}\omega_z^{*}$ & $[\qty{-1.0}{\radian/\second},\qty{1.0}{\radian/\second}]$ \\
            Velocity command resampling range & $[\qty{10.0}{\second},\qty{15.0}{\second}]$ \\
        \midrule
            \multicolumn{2}{c}{Network architecture} \\
        \midrule
            Actor MLP dimensions & $(128,64,32)$ \\
            Critic MLP dimensions & $(64,64)$ \\
            Activation function & ELU \\
        \bottomrule
    \end{tabular}
\end{center}
\end{table}

\end{document}

%% file: acronyms.tex
\acrodef{AD}{Automatic Differentiation}
\acrodef{BPTT}{Backpropagation Through Time}
\acrodef{CRBA}{Composite-Rigid-Body Algorithm}
\acrodef{DoF}{Degree of Freedom}
\acrodef{EoM}{Equation of Motion}
\acrodef{FoG}{First-order Gradient}
\acrodef{GD}{Gradient Descent}
\acrodef{LCP}{Linear Complementarity Problem}
\acrodef{LM}{Levenberg-Marquardt}
\acrodef{LSTM}{Long Short-Term Memory}
\acrodef{NCP}{Nonlinear Complementarity Problem}
\acrodef{PPO}{Proximal Policy Optimization}
\acrodef{RNEA}{Recursive Newton-Euler Algorithm}
\acrodef{RL}{Reinforcement Learning}
\acrodef{RSL}{Robotic Systems Lab}
\acrodef{SHAC}{Short-Horizon Actor-Critic}
\acrodef{ToI}{Time of Impact}
\acrodef{ZoG}{Zeroth-order Gradient}
\acrodef{PD}{Proportional-Derivative}
\acrodef{IFT}{Implicit Function Theorem}